\documentclass[a4paper,fleqn]{cas-dc}

\usepackage[numbers]{natbib}
\usepackage{times}
\usepackage{epsfig}
\usepackage{graphicx}
\usepackage{amsmath}
\usepackage{amssymb}
\usepackage{xcolor}
\usepackage{multirow}
\usepackage{booktabs}
\usepackage{algorithm}
\usepackage{algorithmic}
\usepackage{bm}
\definecolor{highlight}{rgb}{0,0,0}

\definecolor{mygray}{gray}{.6}
\definecolor{myblue}{RGB}{89,158,254}

\def\tsc#1{\csdef{#1}{\textsc{\lowercase{#1}}\xspace}}
\tsc{WGM}
\tsc{QE}
\tsc{EP}
\tsc{PMS}
\tsc{BEC}
\tsc{DE}
\begin{document}
\let\WriteBookmarks\relax
\def\floatpagepagefraction{1}
\def\textpagefraction{.001}
\shorttitle{Neurocomputing}
\shortauthors{S. Zhou et~al.}  
\title[mode = title]{SAVTrack: Selective Vote Aggregation for Reliability-Aware Point Cloud Tracking}

\author[1,2]{Sifan Zhou$^\dagger$$^*$}
\author[3]{Linyue Tan$^\dagger$}
\author[4]{Qiwei Wang$^\dagger$}
\author[1,2]{Ziyu Zhao}

\author[1,2]{Xiaobo Lu\corref{cor1}}
\cortext[cor1]{Corresponding author: Sifan Zhou, sifanjay@gmail.com; Xiaobo Lu, xblu2013@126.com.}

\address[1]{School of Automation, Southeast University, Nanjing, China}
\address[2]{Key Laboratory of Measurement and Control of Complex Systems of Engineering, Ministry of Education, Nanjing, China}
\address[3]{University of Pennsylvania, Philadelphia, PA, USA}
\address[4]{Harbin Institute of Technology (Shenzhen), Shenzhen, China}

\begin{abstract}
    3D single object tracking (SOT) in LiDAR point clouds is essential for autonomous systems, but remains challenging under sparse and incomplete observations. In such cases, different target points provide highly uneven constraints on the object center, causing some point-to-center votes to be substantially less reliable than others. Existing point-based trackers typically aggregate these hypotheses without explicitly modeling their reliability, allowing inaccurate votes to contaminate proposal clustering and degrade localization accuracy. To address this issue, we propose \textbf{SAVTrack}, a motion-aware tracking framework with \textbf{Selective Vote Aggregation (SAV)}. SAVTrack estimates the reliability of each candidate vote from both local seed features and inter-frame motion context, and removes low-confidence hypotheses before proposal clustering. This pre-aggregation gating prevents unreliable hypotheses from affecting cluster formation while introducing only modest computational overhead. SAVTrack achieves competitive performance on KITTI and nuScenes, reaching 68.4/87.4 and 58.44/69.82 Success/Precision, respectively, while running at 82 FPS. It retains fewer than one-sixth of the candidate votes used by dense aggregation and remains particularly effective under sparse target observations.
\end{abstract}


\maketitle
\section{Introduction}

\begin{figure*}[!htbp]
\centering
\includegraphics[width=0.95\linewidth]{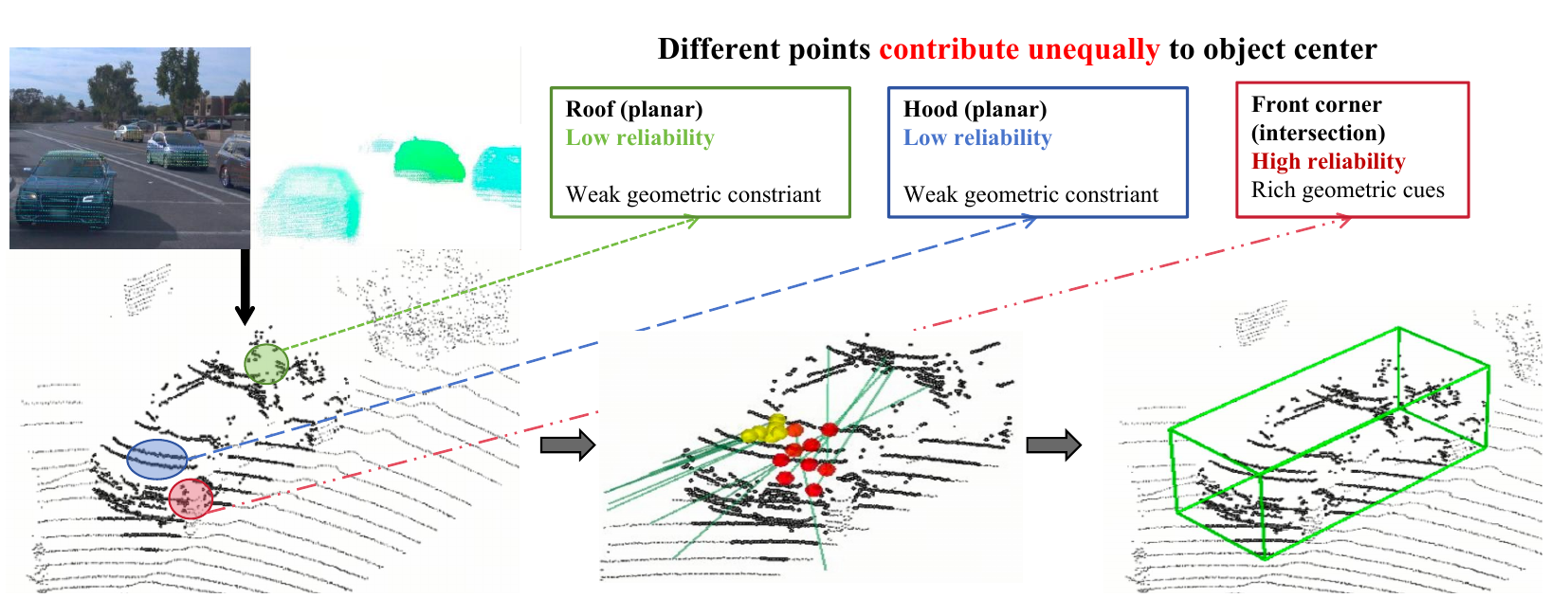}
\caption{Illustration of heterogeneous vote reliability across objects based on point clouds. Points on locally planar or ambiguous regions often produce noisy center hypotheses due to weak geometric constraint, while structurally distinctive regions provide more reliable localization and geometric cues. Filtering low-confidence votes before clustering leads to a more concentrated vote distribution and more accurate proposal formation.}
\label{fig:teaser}
\vspace{-3mm}
\end{figure*}

LiDAR point clouds have wide applications in various computer vision tasks~\cite{zhou2023fastpillars,zhou2025pillarhist,wang2025point4bit,zhou2024lidarptq,ascnet,mspefusion}. Among them, 3D Single Object Tracking (SOT) in LiDAR point clouds is a fundamental perception task for autonomous systems and robotics~\cite{cui2019point,pillartrack,beyond,zhou2026comptrack,trackingreview,cftrack,hu2026tftrack}, requiring accurate target localization across frames under sparse observations, occlusion, and background clutter. Existing 3D SOT approaches can be broadly divided into two paradigms. Siamese-based appearance matching trackers~\cite{p2b,ptt,ptt-journal} localize the target by matching template and search features, benefiting from explicit target appearance correspondence but often becoming vulnerable when point observations are sparse, incomplete, or contaminated by distractors. Motion-centric trackers~\cite{m2track,m2track++,p2p}, in contrast, estimate the target state from inter-frame motion cues, reducing their dependence on appearance matching and offering an effective alternative under appearance ambiguity. Orthogonal to this paradigm distinction, point-based trackers can preserve fine-grained local evidence throughout localization rather than directly collapsing the entire target observation into a single prediction. In particular, point-wise voting allows individual seed\footnote{A seed denotes a sampled point together with its associated point-wise feature, which serves as an elementary voting unit for generating candidate target-center hypotheses.} points to cast explicit hypotheses toward the target center, making the contribution of each local observation directly accessible during proposal generation. This fine-grained formulation provides a natural basis for exploiting local geometric evidence in sparse LiDAR tracking, while also exposing an important question: \emph{do all point-to-center hypotheses contribute equally reliable localization cues?}

In practice, point-to-center hypotheses can exhibit substantially different localization reliability. As shown in Fig.~\ref{fig:teaser}, under sparse and incomplete observations, seed points from different target regions provide unequal geometric constraints on the object center. Specifically, points on large and locally planar surfaces often offer weak geometric cues, which can lead to ambiguous center estimates, whereas structurally distinctive regions, such as surface intersections or corner areas with larger local shape variation, tend to provide stronger localization evidence. Similar observations have also been reported in 3D object detection~\cite{du2020spot,nivssd}. However, this reliability heterogeneity is largely ignored in conventional voting-based pipelines, which typically pass all seed-generated hypotheses to the subsequent clustering stage without explicitly distinguishing their quality. Consequently, inaccurate votes from ambiguous regions can interfere with neighborhood construction and proposal formation, ultimately degrading target localization. This observation motivates us to explicitly model seed-level vote reliability before proposal aggregation.

Although recent 3D tracking methods have substantially advanced feature representation, target correspondence, and motion estimation~\cite{stnet,glt,cutrack,m2track,p2p,focustrack}, the reliability of individual seed-to-center hypotheses before proposal aggregation remains comparatively underexplored. Selective voting has been studied in the related setting of single-frame 3D object detection. In particular, SPOT~\cite{du2020spot} demonstrates that point-to-center hypotheses exhibit heterogeneous localization quality and improves proposal generation by suppressing unreliable votes. 3D SOT, however, introduces a different localization setting: the target is specified across consecutive frames, and the quality of a candidate vote can depend on both local geometric evidence and the temporal motion context of the tracked object. This distinction suggests that vote reliability in 3D tracking should be considered in a target-conditioned temporal context rather than solely from static local geometry~\cite{du2020spot}. However, how to explicitly estimate and exploit such reliability before proposal aggregation remains largely unexplored in existing point-cloud tracking frameworks.

To address this gap, we propose \textbf{SAVTrack}, a motion-centric 3D tracking framework built around \textbf{Selective Vote Aggregation (SAV)}. SAV predicts a per-seed posterior over predefined center-relative sub-regions from fused point-wise and inter-frame motion representations. This posterior serves two purposes: \emph{(i)} it provides a directly supervised estimate of the center-relative region for each seed; and \emph{(ii)} the posterior probability associated with each candidate provides
a confidence score for filtering unreliable votes before proposal clustering. Unlike soft posterior weighting, which retains all hypotheses during aggregation, SAV applies hard pre-aggregation gating to remove low-confidence votes from the candidate set, preventing unreliable hypotheses from affecting local neighborhood construction and subsequent proposal formation. The resulting module is lightweight, consisting of a small MLP classifier and region-specific vote regressors, and preserves real-time inference. We further validate the learned confidence through aggregation-strategy comparisons and geometry--reliability analysis in Sec.~\ref{sec:exp}. Overall, our contributions are as follows:

\begin{itemize}
    \item We identify and systematically study seed-level vote reliability in 3D single object tracking, showing that point-to-center hypotheses exhibit substantial reliability differences with respect to local geometry and observation sparsity.

    \item We propose \textbf{Selective Vote Aggregation (SAV)}, a lightweight motion-conditioned gating mechanism that estimates vote reliability from joint point-wise and inter-frame motion representations.

    \item Extensive experiments on two widely used benchmarks, KITTI and nuScenes, demonstrate that SAVTrack achieves competitive tracking accuracy at 82~FPS while retaining fewer than one-sixth of the candidate votes used by dense aggregation.
\end{itemize}

\begin{figure*}[!htbp]
\centering
\includegraphics[width=\linewidth]{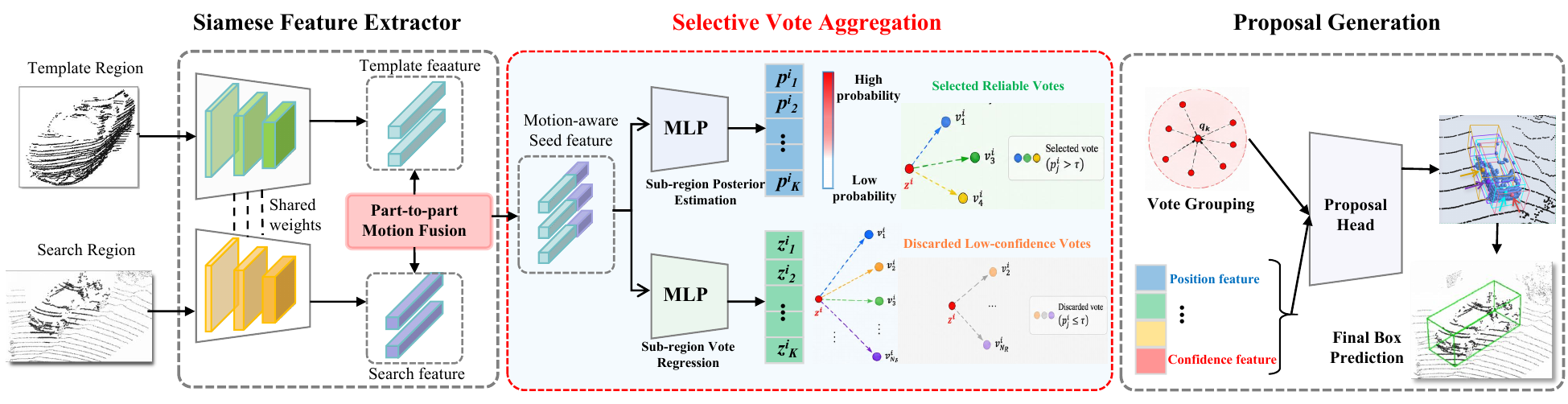}
\caption{Overview of SAVTrack. Given the target observation from the previous frame and the current search region, a weight-shared PointNet++~\cite{pointnet++} backbone extracts point-wise features, which are fused through the P2P-derived part-to-part motion representation to construct motion-aware seed features. The proposed Selective Vote Aggregation (SAV) module jointly performs center-relative sub-region posterior estimation and region-specific vote regression to generate multiple candidate center hypotheses for each seed. The posterior probability associated with each candidate is used as a proxy for vote reliability, and low-confidence hypotheses are removed by hard pre-aggregation gating before proposal formation. The retained reliable votes are then grouped and aggregated to generate target proposals, from which the final 3D target box is predicted in an end-to-end manner.}
\label{fig:framework}
\vspace{-4mm}
\end{figure*}

\vspace{-2mm}
\section{Related Work}

\noindent\textbf{Siamese-based 3D Single Object Tracking.}
Early 3D SOT methods formulate tracking as a template--search appearance matching problem using Siamese networks. As a pioneer, SC3D~\cite{sc3d} introduces the first Siamese framework that extracts features from a template and candidate regions and selects the best match via feature distance. Subsequently, 3D-SiamRPN~\cite{3dsiamrpn} extends this paradigm with a region proposal network for 3D localization. P2B~\cite{p2b} integrates VoteNet-style point-wise voting into the Siamese framework to generate target proposals in an end-to-end manner, inspiring a series of follow-up works~\cite{ptt,bat,lttr,cmt,stnet,osp2b,glt,hu2025mvctrack,pillartrack}. For instance, BAT~\cite{bat} encodes object size priors to augment template--search correlation; PTT~\cite{ptt,ptt-journal}, LTTR~\cite{lttr}, CMT~\cite{cmt}, and STNet~\cite{stnet} explore transformer- or attention-based feature propagation and correlation; CXTrack~\cite{cxtrack} emphasizes contextual modeling with a target-centric transformer; and MBPTrack~\cite{mbptrack} introduces an external memory to enhance spatial--temporal information aggregation. OSP2B~\cite{osp2b} revisits the point-to-box localization stage and replaces the conventional two-stage proposal generation and scoring process with a one-stage formulation, jointly predicting 3D proposals and center-ness scores while introducing a target classifier to suppress interference proposals. Despite these advances, existing Siamese trackers mainly improve template--search correspondence, feature interaction, or proposal-level scoring, while the reliability of individual seed-to-center hypotheses before vote aggregation remains largely unmodeled. Moreover, sparse and incomplete point observations continue to pose challenges to appearance-based matching.

\noindent\textbf{Motion-based 3D Single Object Tracking.} Motion-based trackers formulate 3D SOT as inter-frame motion estimation rather than template--search appearance matching. M$^2$Track~\cite{m2track} first segments foreground points in the search region and then infers the target's 4-DOF relative motion offset; M$^2$Track++~\cite{m2track++} further extends this paradigm to semi-supervised settings. P2P~\cite{p2p} demonstrates the effectiveness of a motion-centric formulation by transforming appearance matching into part-to-part relative offset estimation between adjacent frames. More recent works explore additional motion cues: VoxelTrack~\cite{voxeltrack} exploits multi-level voxel representations for 3D spatial reasoning; DMT~\cite{dmt} introduces a motion prediction module that estimates the target center from historical bounding boxes and refines it with point features. These approaches improve foreground motion modeling, whereas SAV addresses a complementary question in our voting formulation: whether a candidate seed-to-center hypothesis should enter clustering at all. SAV therefore estimates a directly supervised per-vote posterior and applies the decision before proposal clustering.

\section{Method}
\label{sec:method}

\subsection{3D SOT Task Definition}
Given an initial target specified by its ground-truth 3D bounding box in the first frame, 3D SOT aims to sequentially estimate the target state in each subsequent LiDAR frame. 
At frame $t$, following the standard tracking-by-cropping protocol~\cite{ptt-journal,m2track,p2p}, a search region is extracted from the current LiDAR frame around the target location estimated in the previous frame. Specifically, the previous prediction $\mathbf{B}_{t-1}$ is spatially enlarged by a predefined margin to form a search crop, from which the current search point cloud
$\mathcal{P}_{t}=\{\mathbf{p}_{j}^{t}=[x_j,y_j,z_j,r_j]^\top\in\mathbb{R}^{4}\}$
is obtained. The corresponding target observation from the previous frame is denoted as
$\mathcal{P}_{t-1}=\{\mathbf{p}_{i}^{t-1}=[x_i,y_i,z_i,r_i]^\top\in\mathbb{R}^{4}\}$,
where $(x,y,z)$ are the point coordinates and $r$ denotes the LiDAR reflectance. Given the template point cloud $\mathcal{P}_{t-1}$ and search point cloud $\mathcal{P}_{t}$, the tracker predicts the current 3D bounding box
$\mathbf{B}_{t}=[x,y,z,h,w,l,\theta]^\top\in\mathbb{R}^{7}$,
where $(x,y,z)$ denote the box center, $(h,w,l)$ represent its dimensions, and $\theta$ is the heading angle. Following common practice in 3D SOT~\cite{p2b,ptt,bat,p2p}, the target size is assumed to remain unchanged throughout a sequence. Therefore, tracking reduces to estimating the target center and heading angle at each frame.

\subsection{SAVTrack Framework Overview}
\label{sec:framework}

Fig.~\ref{fig:framework} illustrates the overall architecture of proposed \textbf{SAVTrack}. We adopt P2P-point~\cite{p2p} as our motion-centric baseline, which models inter-frame target dynamics through part-to-part feature interaction between consecutive point clouds. Its learned motion representation captures target displacement across frames and provides the temporal context for current-frame localization. To enable point-wise proposal generation, we employ a standard voting-based localization head on top of the P2P-point~\cite{p2p} representation. Our proposed \textbf{Selective Vote Aggregation (SAV)} is then introduced before proposal clustering to explicitly estimate candidate-vote reliability and suppress unreliable center hypotheses.

\begin{figure*}[t]
\centering
\includegraphics[width=0.95\textwidth]{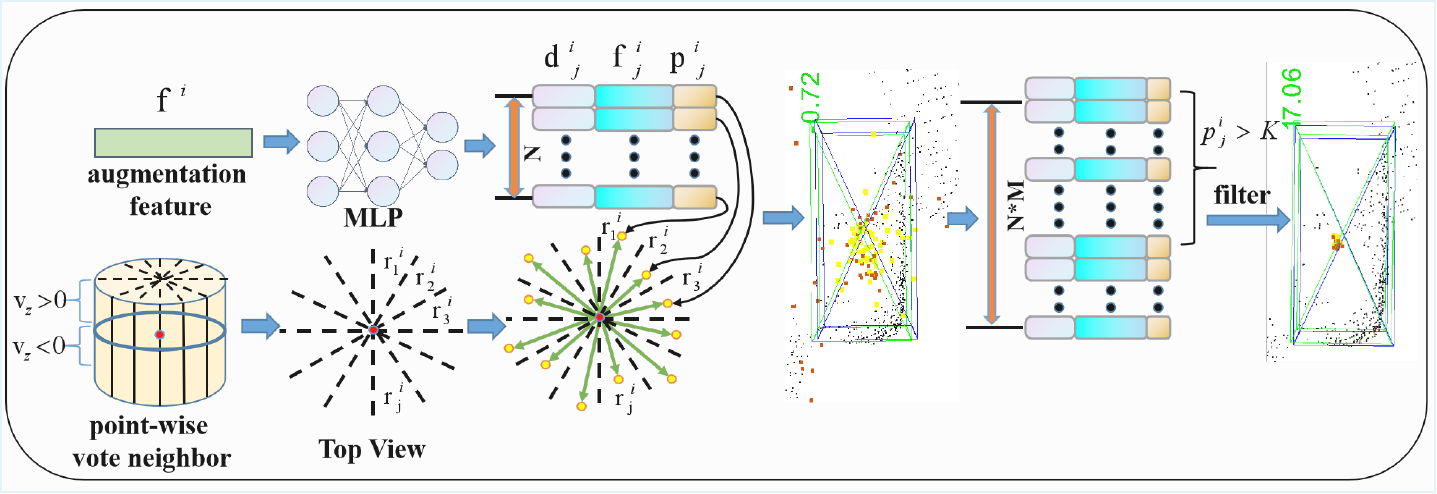}
\vspace{-2mm}
\caption{Illustration of the Selective Vote Aggregation (SAV) module. For each seed point, SAV partitions the center-relative space into $N_R$ pre-defined sub-regions and learns a posterior distribution over which sub-region contains the true object center. The posterior confidence is used as a proxy for candidate-vote reliability: votes with high posterior probability are retained, while low-confidence votes are excluded before proposal clustering.}
\label{fig:SAV}
\vspace{-4mm}
\end{figure*}

\noindent\textbf{Motion-aware Seed Representation.}
Given two cropped point clouds from consecutive frames,
$\mathcal{P}_{t-1}^{\mathrm{crop}}$ and $\mathcal{P}_{t}^{\mathrm{crop}}$,
a weight-shared PointNet++~\cite{pointnet++} encoder extracts point-wise features
$\mathbf{f}_{t-1},\mathbf{f}_{t}\in\mathbb{R}^{N\times C}$.
Following the part-to-part motion modeling of P2P~\cite{p2p},
the features from the two frames are globally aggregated and fused to obtain an inter-frame motion representation
$\mathbf{F}_{\mathrm{m}}$ that characterizes the target dynamics between adjacent observations. 

We sample $N_s$ seed points from the current-frame point cloud using farthest point sampling (FPS) and retain their corresponding point-wise features.
The motion representation $\mathbf{F}_{\mathrm{m}}$ is then broadcast to all seeds and concatenated with each seed feature:
\vspace{-2mm}
\begin{equation}
\begin{aligned}
    \tilde{\mathbf{f}}^{\,i}
    =
    [\mathbf{f}^{\,i};\mathbf{F}_{\mathrm{m}}],
    \qquad i=1,\ldots,N_s ,
\end{aligned}
\end{equation}
where $\tilde{\mathbf{f}}^{\,i}$ denotes the motion-aware representation of the $i$-th seed.
This representation preserves local point-wise evidence while incorporating temporal target dynamics, and is subsequently used by SAV for candidate-vote confidence estimation and center hypothesis generation.

\subsection{Selective Vote Aggregation}
\label{sec:sav}

Building on the motion-aware seed representation, we introduce
\textbf{Selective Vote Aggregation (SAV)} to explicitly distinguish candidate
center hypotheses before proposal clustering.
Selective voting has previously been explored in single-frame 3D object detection~\cite{du2020spot};
in contrast, SAV estimates candidate confidence from a target-conditioned representation
that jointly incorporates current-frame point-wise evidence and inter-frame motion context.
Specifically, each seed predicts a posterior over predefined center-relative sub-regions
and generates a corresponding set of candidate center hypotheses.
The posterior confidence is then used for hard pre-aggregation gating, such that
low-confidence hypotheses are removed before participating in proposal clustering.

\noindent\textbf{Center-relative Sub-region Posterior Estimation.}
For each seed $i$, we consider the relative displacement between its coordinate
$\mathbf{z}^{i}$ and the ground-truth target center $\mathbf{c}$.
The center-relative displacement space is partitioned into $N_R$ predefined
sub-regions $\{r_j\}_{j=1}^{N_R}$, as illustrated in Fig.~\ref{fig:SAV}.
Accordingly, the ground-truth region label of seed $i$ is defined as
\vspace{-2mm}
\begin{equation}
\begin{aligned}
    y^{i}
    =
    \mathcal{Q}
    \left(
    \mathbf{c}-\mathbf{z}^{i}
    \right),
    \qquad
    y^{i}\in\{1,\ldots,N_R\},
\end{aligned}
\end{equation}
where $\mathcal{Q}(\cdot)$ denotes the predefined region-assignment operator.

Given the motion-aware seed feature $\tilde{\mathbf{f}}^{\,i}$,
a lightweight classifier $g(\cdot;\theta^{g})$ predicts a categorical posterior
over the $N_R$ center-relative sub-regions:
\begin{equation}
\begin{aligned}
    \mathbf{p}^{i}
    &=
    \operatorname{Softmax}
    \left(
    g(\tilde{\mathbf{f}}^{\,i};\theta^{g})
    \right),
\end{aligned}
\end{equation}
where
\begin{equation}
\begin{aligned}
    p_j^{i}
    =
    P
    \left(
    y^{i}=j
    \mid
    \tilde{\mathbf{f}}^{\,i}
    \right)
\end{aligned}
\end{equation}
denotes the posterior probability that the target center belongs to sub-region $j$ relative to seed $i$. The classifier is supervised by the ground-truth sub-region labels using
cross-entropy loss:
\vspace{-2mm}
\begin{equation}
\begin{aligned}
    L_{\mathrm{vote\text{-}cls}}
    =
    -\frac{1}{N_s}
    \sum_{i=1}^{N_s}
    \log p_{y^{i}}^{i}.
\end{aligned}
\end{equation}
Note that the classifier is supervised to predict the center-relative sub-region,
rather than an explicit reliability label.
The resulting posterior can nevertheless provide a confidence measure for each
candidate hypothesis, reflecting how strongly the learned motion-aware representation
supports the corresponding center-relative configuration.
We therefore use the posterior confidence as a proxy for vote reliability during
aggregation.

\noindent\textbf{Vote Regression and Reliability Gating.}
Concurrent with confidence estimation, each seed point generates candidate center votes for all $N_R$ sub-regions. Specifically, for each sub-region $j$, a region-specific MLP regressor $\phi_j(\cdot; \theta_j^\phi)$ takes the seed feature $\tilde{\mathbf{f}}^i$ and predicts the offset from $\mathbf{z}^i$ to the object center:
\vspace{-2mm}
\begin{equation}
\begin{aligned}
    \Delta\mathbf{z}_j^i = \phi_j(\tilde{\mathbf{f}}^i; \theta_j^\phi), \quad
    \hat{\mathbf{d}}_j^i = \mathbf{z}^i + \Delta\mathbf{z}_j^i,
\end{aligned}
\end{equation}
yielding a predicted center location $\hat{\mathbf{d}}_j^i$. The regressor is supervised by the ground-truth offset $\Delta\mathbf{z}^i_* = \mathbf{c} - \mathbf{z}^i$ using an $L_1$ loss applied only to the correct sub-region:

\vspace{-2mm}
\begin{equation}
\begin{aligned}
    L_{\text{vote-reg}} = \sum_{j=1}^{N_R} \sum_{i: y^i=j} \|\phi_j(\tilde{\mathbf{f}}^i; \theta_j^\phi) - \Delta\mathbf{z}^i_*\|_1.
\end{aligned}
\end{equation}

At inference time, each seed generates $N_R$ candidate center hypotheses,
$\{v_j^i\}_{j=1}^{N_R}$, where
$v_j^i=(\hat{\mathbf d}_j^i,p_j^i)$.
The posterior $p_j^i$ provides a confidence measure for the corresponding
candidate and is used as a proxy for vote reliability.
SAV performs hard pre-aggregation gating:
\vspace{-2mm}
\begin{equation}
\begin{aligned}
    \mathcal V_{\mathrm{SAV}}
    =
    \left\{
    v_j^i \mid p_j^i > \tau
    \right\}.
\end{aligned}
\end{equation}
Unlike confidence reweighting, which preserves all hypotheses in the
aggregation set, hard gating changes the candidate set itself and prevents
discarded votes from affecting neighborhood construction during proposal
clustering.

The total vote loss is:
\begin{equation}
\begin{aligned}
    L_{\text{vote}} = L_{\text{vote-cls}} + \lambda_{\text{reg}} L_{\text{vote-reg}},
\end{aligned}
\end{equation}
where $\lambda_{\text{reg}}$ is a balancing hyper-parameter. Algorithm~\ref{alg:savtrack} summarizes the complete inference pipeline of SAVTrack.

\begin{algorithm}[!htbp]
\caption{SAVTrack Inference Pipeline}
\label{alg:savtrack}
\textbf{Input:} Template crop $\mathcal{P}_{t-1}^{\text{crop}}$, search crop $\mathcal{P}_t^{\text{crop}}$, previous box $\mathbf{B}_{t-1}$; hyper-parameters: $N_R$, $\tau$, $N_{\text{prop}}$, $r_p$ \\
\textbf{Output:} Predicted bounding box $\mathbf{B}_t$
\begin{algorithmic}[1]
\vspace{1mm}
\STATE \textbf{// Feature encoding}
\STATE Extract per-point features $\mathbf{f}_{t-1}, \mathbf{f}_t$ via weight-shared PointNet++
\STATE Compute global features $\mathbf{F}_{t-1}, \mathbf{F}_t$ via max-pooling
\STATE Fuse motion feature $\mathbf{F}_{\text{pp}}^{\text{fusion}}$ via cascaded 1D convolutions
\STATE Sample $N_s$ seed points via FPS; obtain motion-aware seed features $\tilde{\mathbf{f}}^i$
\vspace{1mm}
\STATE \textbf{// Selective Vote Aggregation (SAV)}
\STATE Predict sub-region posteriors $[p_1^i, \dots, p_{N_R}^i] = g(\tilde{\mathbf{f}}^i)$ for each seed
\STATE Regress per-region vote offsets $\Delta\mathbf{z}_j^i = \phi_j(\tilde{\mathbf{f}}^i)$ and vote centers $\hat{\mathbf{d}}_j^i$
\STATE Collect filtered votes $\mathcal{V}_{\text{filtered}} = \{v_j^i : p_j^i > \tau\}$ \hfill \COMMENT{Hard reliability gating}
\IF{$|\mathcal{V}_{\text{filtered}}| = 0$}
    \STATE Fall back to top-$K_0$ posterior-ranked votes before thresholding \hfill \COMMENT{Empty-vote fallback}
\ENDIF
\vspace{1mm}
\STATE \textbf{// Proposal generation and tracking}
\STATE Select $K = \min(N_{\text{prop}}, |\mathcal{V}_{\text{filtered}}|)$ cluster centroids from $\mathcal{V}_{\text{filtered}}$
\FOR{each centroid}
    \STATE Gather neighboring votes within radius $r_p$ via ball query
    \STATE Predict proposal box $\mathbf{b}_k$ and objectness score $s_k$ via MLP $\Theta$
\ENDFOR
\RETURN $\mathbf{B}_t$
\end{algorithmic}
\end{algorithm}

\subsection{Proposal Generation and Training}
\label{sec:proposal_generation}

Given the filtered high-confidence vote set
$\mathcal{V}_{\mathrm{SAV}}$, we generate target proposals through
vote clustering and proposal refinement.
Let $M=|\mathcal{V}_{\mathrm{SAV}}|$ denote the number of retained
votes after selective gating.
Following the VoteNet-style aggregation paradigm~\cite{votenet},
a set of representative vote centers is first sampled from
$\mathcal{V}_{\mathrm{SAV}}$ as proposal seeds.
For each proposal seed, neighboring votes within a predefined spatial
radius are grouped by ball query to construct a local vote cluster:
\begin{equation}
\begin{aligned}
    \mathcal{T}_k
    =
    \left\{
    v_j^i \in \mathcal{V}_{\mathrm{SAV}}
    \;\middle|\;
    \left\|
    \hat{\mathbf d}_j^i-\mathbf q_k
    \right\|_2
    < r_p
    \right\},
\end{aligned}
\end{equation}
where $\mathbf q_k$ denotes the center of the $k$-th proposal seed and
$r_p$ is the grouping radius.
Since unreliable hypotheses have already been removed by SAV, the resulting
clusters are constructed only from the retained high-confidence center
hypotheses, thereby reducing the influence of ambiguous votes on local
proposal formation.

For each vote cluster $\mathcal{T}_k$, the spatial coordinates and associated
vote features are aggregated to form a cluster-level representation.
A lightweight proposal head $\Theta(\cdot)$ then predicts the target state
$\mathbf b_k$ with an objectness score $s_k$:
\begin{equation}
\begin{aligned}
    (\mathbf b_k,s_k)
    =
    \Theta(\mathcal{T}_k).
\end{aligned}
\end{equation}
The proposal state is parameterized as:
\begin{equation}
\begin{aligned}
    \mathbf b_k
    =
    [x_k,y_k,z_k,\theta_k]^\top,
\end{aligned}
\end{equation}
where $(x_k,y_k,z_k)$ denote the predicted target center and $\theta_k$
is the heading angle.
Following the standard 3D SOT setting, the target dimensions are inherited
from the initialized bounding box and remain fixed throughout the sequence.
The objectness score $s_k$ measures the likelihood that the corresponding
proposal accurately represents the tracked target.
After all candidate proposals are generated, the proposal with the highest
valid objectness score is selected as the predicted target state
$\mathbf B_t$ for the current frame.

\noindent\textbf{Proposal Supervision.}
Following VoteNet-style proposal supervision~\cite{votenet}, proposal candidates are assigned positive or negative labels according to their center distance to the ground-truth target center. The objectness score $s_k$ is optimized using a binary classification loss $L_{\mathrm{prop}}$, while the target state of positive proposals is supervised by a regression loss $L_{\mathrm{box}}$.

Combining the selective voting objective introduced in Sec.~\ref{sec:sav}, the overall training objective of SAVTrack is
\begin{equation}
\begin{aligned}
    L
    =
    L_{\mathrm{vote}}
    +
    \gamma_1 L_{\mathrm{prop}}
    +
    \gamma_2 L_{\mathrm{box}},
\end{aligned}
\end{equation}
where $\gamma_1$ and $\gamma_2$ balance proposal classification and box regression, respectively. All components are optimized jointly in an end-to-end manner.

\section{Experiments}
\label{sec:exp}
\begin{table*}[!t]
\caption{Comparison with state-of-the-art methods on the KITTI dataset~\cite{kitti}.
\textit{Success} / \textit{Precision} are reported for evaluation.
\textbf{Bold} and \underline{underlined} values denote the best and second-best results, respectively. Rep means the points' view representation.}
\centering
\resizebox{\textwidth}{!}{
\normalsize
\begin{tabular}{c|l|c|c|c|cccc|cc}
\toprule[0.4mm]
\multirow{2}{*}{Paradigm}
& \multirow{2}{*}{Tracker}
& \multirow{2}{*}{Source}
& \multirow{2}{*}{Rep.}
& Mean
& Car
& Pedestrian
& Van
& Cyclist
& \multirow{2}{*}{FPS}
& \multirow{2}{*}{Device} \\
&&&& (14,068) & (6,424) & (6,088) & (1,248) & (308) && \\
\midrule

\multirow{15}{*}{Siamese}
& SC3D~\cite{sc3d} & CVPR'19 & Point
& 31.2 / 48.5
& 41.3 / 57.9
& 18.2 / 37.8
& 40.4 / 47.0
& 41.5 / 70.4
& 2 & GTX 1080Ti \\

& P2B~\cite{p2b} & CVPR'20 & Point
& 42.4 / 60.0
& 56.2 / 72.8
& 28.7 / 49.6
& 40.8 / 48.4
& 32.1 / 44.7
& 40 & GTX 1080Ti \\

& 3D-SiamRPN~\cite{3dsiamrpn} & IEEE Sensors J & Point
& 46.6 / 64.9
& 58.2 / 76.2
& 35.2 / 56.2
& 45.7 / 52.9
& 36.2 / 49.0
& 45 & GTX 1080Ti \\

& PTT~\cite{ptt} & IROS'21 & Point
& 55.1 / 74.2
& 67.8 / 81.8
& 44.9 / 72.0
& 43.6 / 52.5
& 37.2 / 47.3
& 40 & GTX 1080Ti \\

& LTTR~\cite{lttr} & BMVC'21 & BEV
& 48.7 / 65.8
& 65.0 / 77.1
& 33.2 / 56.8
& 35.8 / 45.6
& 66.2 / 89.9
& 23 & GTX 1080Ti \\

& BAT~\cite{bat} & ICCV'21 & Point
& 51.2 / 72.8
& 60.5 / 77.7
& 42.1 / 70.1
& 52.4 / 67.0
& 33.7 / 45.4
& 57 & RTX 2080 \\

& V2B~\cite{v2b} & NeurIPS'21 & Point+BEV
& 58.4 / 75.2
& 70.5 / 81.3
& 48.3 / 73.5
& 50.1 / 58.0
& 40.8 / 49.7
& 37 & TITAN RTX \\

& PTTR~\cite{pttr} & CVPR'22 & Point
& 57.9 / 78.2
& 65.2 / 77.4
& 50.9 / 81.6
& 52.5 / 61.8
& 65.1 / 90.5
& 50 & Tesla V100 \\

& STNet~\cite{stnet} & ECCV'22 & Point
& 61.3 / 80.1
& 72.1 / 84.0
& 49.9 / 77.2
& 58.0 / 70.6
& 73.5 / 93.7
& 35 & TITAN RTX \\

& GLT-T~\cite{glt} & AAAI'23 & Point
& 60.1 / 79.3
& 68.2 / 82.1
& 52.4 / 78.8
& 52.6 / 62.9
& 68.9 / 92.1
& 30 & GTX 1080Ti \\

& OSP2B~\cite{osp2b} & IJCAI'23 & Point
& 60.5 / 82.3
& 67.5 / 82.3
& 53.6 / 85.1
& 56.3 / 66.2
& 65.6 / 90.5
& 34 & GTX 1080Ti \\

& CXTrack~\cite{cxtrack} & CVPR'23 & Point
& 67.5 / 85.3
& 69.1 / 81.6
& 67.0 / 91.5
& 60.0 / 71.8
& 74.2 / 94.3
& 34 & RTX 3090 \\


& SyncTrack~\cite{synctrack} & ICCV'23 & Point
& 64.1 / 81.9
& 73.3 / 85.0
& 54.7 / 80.5
& 60.3 / 70.0
& 73.1 / 93.8
& 45 & TITAN RTX \\

& MoCUT~\cite{cutrack} & ICLR'24 & Point
& 65.8 / 85.0
& 67.6 / 80.5
& 63.3 / 90.0
& 64.5 / 78.8
& \textbf{76.7} / 94.2
& 48 & RTX 3070Ti \\

& CFTrack~\cite{cftrack} & Neurocomputing'26 & Point
& 59.7 / 79.3
& 71.4 / 83.2
& 49.8 / 80.0
& 52.1 / 57.9
& 43.9 / 54.5
& 45 & RTX 4090 \\

\midrule

\multirow{7}{*}{Motion}
& M$^2$Track~\cite{m2track} & CVPR'22 & Point
& 62.9 / 83.4
& 65.5 / 80.8
& 61.5 / 88.2
& 53.8 / 70.7
& 73.2 / 93.5
& 57 & Tesla V100 \\

& M$^2$Track++~\cite{m2track++} & TPAMI'23 & Point
& 66.5 / 85.2
& 71.1 / 82.7
& 61.8 / 88.7
& 62.8 / 78.5
& 75.9 / 94.0
& 57 & Tesla V100 \\

& VoxelTrack~\cite{voxeltrack} & ACM MM'24 & Voxel
& 70.4 / 88.3
& 72.5 / 84.7
& 67.8 / 92.6
& \underline{69.8} / \underline{83.6}
& 75.1 / \underline{94.7}
& 36 & TITAN RTX \\

& FocusTrack~\cite{focustrack} & ACM MM'25 & BEV
& 71.3 / \textbf{89.4}
& \textbf{74.1} / \textbf{85.9}
& 69.3 / \underline{94.1}
& 68.4 / 83.5
& 75.9 / \underline{94.7}
& 105 & RTX 3090 \\

& CompTrack~\cite{zhou2026comptrack} & AAAI'26 & BEV
& \underline{71.4} / \underline{89.3}
& 73.4 / 85.2
& \underline{69.5} / \textbf{94.7}
& 68.5 / 82.5
& \underline{76.0} / \textbf{94.8}
& 90 & RTX 3090 \\

& \textit{P2P-voxel}~\cite{p2p} & IJCV'25 & Voxel
& \textbf{71.7} / \textbf{89.4}
& \underline{73.6} / \underline{85.7}
& \textbf{69.6} / 94.0
& \textbf{70.3} / \textbf{83.9}
& 75.5 / 94.6
& 71 & RTX 3090 \\
& \textit{P2P-point}~\cite{p2p} & IJCV'25 & Point
& 66.2 / 85.4
& 68.8 / 81.7
& 62.7 / 89.1
& 65.4 / 80.1
& 74.8 / \textbf{94.8}
& 105 & RTX 3090 \\
\midrule
\rowcolor{myblue!18}
& \textbf{SAVTrack (Ours)} & \textbf{Ours} & Point
& 68.4 / 87.4
& 71.1 / 83.9
& 65.1 / 91.8
& 68.0 / 82.5
& 75.9 / \textbf{94.8}
& 82 & RTX 3090 \\

\bottomrule[0.4mm]
\end{tabular}}
\label{tab:KITTI}
\vspace{-2mm}
\end{table*}

\noindent\textbf{Datasets and Evaluation Metrics.}
We evaluate SAVTrack on two widely used 3D single object tracking benchmarks, KITTI~\cite{kitti} and nuScenes~\cite{nuScenes}. KITTI is collected using a 64-beam LiDAR and contains four commonly evaluated categories, including Car, Pedestrian, Van, and Cyclist, while nuScenes is captured with a 32-beam LiDAR and provides substantially sparser target observations across Car, Pedestrian, Truck, Trailer, and Bus categories. We follow the standard one-pass evaluation (OPE) protocol~\cite{otb2013} adopted by previous 3D SOT methods~\cite{p2b,ptt,bat,m2track,p2p}. Tracking performance is evaluated using Success and Precision, where Success measures the overlap between the predicted and ground-truth 3D bounding boxes, while Precision evaluates the center localization error.

\noindent\textbf{Implementation Details.}
All experiments are implemented in PyTorch and trained using the Adam optimizer with an initial learning rate of $1\times10^{-4}$, a batch size of 128, and 30 training epochs. Unless otherwise specified, SAV uses $N_R=12$ center-relative sub-regions and a confidence threshold of $\tau=0.3$. The loss-balancing coefficients are set to $\lambda_{\mathrm{reg}}=1.0$, $\gamma_1=1.0$, and $\gamma_2=1.0$. During proposal generation, at most $N_{\mathrm{prop}}=256$ proposal seeds are used, and neighboring votes are grouped with a radius of $r_p=0.3$\,m. Following the implementation of the voting baseline, proposals whose centers are sufficiently close to the ground-truth center are treated as positive, whereas distant proposals are treated as negative. Specifically, we use 0.3\,m and 0.6\,m as the positive and negative distance thresholds, respectively, and ignore proposals falling between the two thresholds. All ablation studies are conducted using the same P2P-based voting baseline described in Sec.~\ref{sec:framework}, with SAV being the only additional vote-selection component unless otherwise stated. Training is performed on a single NVIDIA RTX 3090 GPU. Runtime measurements for SAVTrack and the corresponding baseline are conducted on the same hardware under an identical inference protocol to ensure a fair efficiency comparison.

\subsection{Quantitative Experiment}

\noindent\textbf{Comparison on KITTI.}
Tab.~\ref{tab:KITTI} compares SAVTrack with representative Siamese-based and
motion-centric 3D single object trackers on the KITTI benchmark.
SAVTrack achieves an average Success/Precision of 68.4\%/87.4\%.
Despite its lightweight point-based representation, SAVTrack maintains competitive
tracking accuracy while operating at 82~FPS.
Notably, it reaches 94.8\% Precision on Cyclist, matching the best result
reported in the table. The representation-wise comparison provides additional context for these results.
Recent voxel- and BEV-based trackers, such as VoxelTrack~\cite{voxeltrack},
FocusTrack~\cite{focustrack}, CompTrack~\cite{zhou2026comptrack}, and
P2P-voxel~\cite{p2p}, achieve higher absolute mean accuracy in several cases.
These methods employ spatially aggregated voxel or BEV representations and
different localization architectures, whereas SAVTrack operates directly on
point-level seed hypotheses.
Accordingly, the objective of SAVTrack is not to replace the underlying point
representation with a heavier spatial encoding, but to improve how explicit
seed-to-center hypotheses are utilized during proposal formation.
Within this point-based setting, SAVTrack remains competitive with recent
trackers and achieves particularly strong results on Van and Cyclist.

Relative to the published P2P-point tracker~\cite{p2p}, SAVTrack improves the
mean Success/Precision from 66.2/85.4 to 68.4/87.4.
The improvements are consistent across the major categories: +2.3/+2.2 on Car,
+2.4/+2.7 on Pedestrian, +2.6/+2.4 on Van, and +1.1/+0.0 on Cyclist.
The relatively larger gains on Pedestrian and Van are consistent with the
motivation of SAV, since sparse or incomplete target observations can produce
more ambiguous point-wise center hypotheses.
By estimating candidate confidence and removing low-confidence votes before
proposal clustering, SAV reduces the influence of unreliable hypotheses without
changing the underlying point-based representation.
The contribution of the proposed aggregation strategy is isolated more
rigorously through the matched dense, soft-weighted, top-1, and hard-gated
ablations in Sec.~\ref{sec:ablation}, while its behavior under different target
point densities is further analyzed in Sec.~\ref{sec:sparsity_analysis}.

\begin{table*}[!t]
\caption{Comparisons with state-of-the-art methods on nuScenes dataset~\cite{nuScenes}. 
\textit{Success} / \textit{Precision} are used for evaluation. 
The best and second-best results are highlighted in \textbf{bold} and \underline{underline}, respectively. $^\dagger$ denotes results reproduced using the official implementation.}
\centering
\resizebox{1.0\textwidth}{!}{
\small
\begin{tabular}{l|c|ccccc}
\toprule[0.4mm]
\multirow{2}{*}{\textbf{Tracker}} 
& \textbf{Mean} 
& \textbf{Car} 
& \textbf{Pedestrian} 
& \textbf{Truck} 
& \textbf{Trailer} 
& \textbf{Bus} \\
& (117,278) 
& (64,159) 
& (33,227) 
& (13,587) 
& (3,352) 
& (2,953) \\
\midrule

SC3D~\cite{sc3d}
& 20.70 / 20.20
& 22.31 / 21.93
& 11.29 / 12.65
& 30.67 / 27.73
& 35.28 / 28.12
& 29.35 / 24.08 \\

P2B~\cite{p2b}
& 36.48 / 45.08
& 38.81 / 43.18
& 28.39 / 52.24
& 42.95 / 41.59
& 48.96 / 40.05
& 32.95 / 27.41 \\

PTT~\cite{ptt}
& 36.33 / 41.72
& 41.22 / 45.26
& 19.33 / 32.03
& 50.23 / 48.56
& 51.70 / 46.50
& 39.40 / 36.70 \\

BAT~\cite{bat}
& 38.10 / 45.71
& 40.73 / 43.29
& 28.83 / 53.32
& 45.34 / 42.58
& 52.59 / 44.89
& 35.44 / 28.01 \\

V2B~\cite{v2b}
& - / -
& 54.40 / 59.70
& 30.10 / 55.40
& 53.70 / 54.50
& 54.90 / 51.44
& - / - \\

PTTR~\cite{pttr}
& 44.50 / 52.07
& 51.89 / 58.61
& 29.90 / 45.09
& 45.30 / 44.74
& 45.87 / 38.36
& 43.14 / 37.74 \\

GLT-T~\cite{glt}
& 44.42 / 54.33
& 48.52 / 54.29
& 31.74 / 56.49
& 52.74 / 51.43
& 57.60 / 52.01
& 44.55 / 40.69 \\

MoCUT~\cite{cutrack}
& 51.19 / 64.63
& 57.32 / 66.01
& 33.47 / 63.12
& 61.75 / 64.38
& 60.90 / 61.84
& 57.39 / 56.07 \\

MBPTrack~\cite{mbptrack}
& 57.48 / 69.88
& 62.47 / 70.41
& 45.32 / 74.03
& 62.18 / 63.31
& 65.14 / 61.33
& 55.41 / 51.76 \\

\hline

M$^2$Track~\cite{m2track}
& 49.23 / 62.73
& 55.85 / 65.09
& 32.10 / 60.92
& 57.36 / 59.54
& 57.61 / 58.26
& 51.39 / 51.44 \\

PTTR++~\cite{pttr++}
& 51.86 / 60.63
& 59.96 / 66.73
& 32.49 / 50.50
& 59.85 / 61.20
& 54.51 / 50.28
& 53.98 / 51.22 \\

STTracker~\cite{sttracker}
& 49.66 / 66.77
& 56.11 / 69.07
& 37.58 / 68.36
& 54.29 / 60.71
& 48.13 / 55.40
& 36.31 / 36.07 \\

SeqTrack3D~\cite{seqtrack3d}
& 55.92 / 68.94
& 62.55 / 71.46
& 39.94 / 68.57
& 60.97 / 63.04
& 68.37 / 61.76
& 54.33 / 53.52 \\

VoxelTrack~\cite{voxeltrack}
& \underline{59.00} / \textbf{71.40}
& 63.90 / 71.60
& \textbf{46.80} / \textbf{75.90}
& \underline{64.80} / \underline{65.90}
& 69.50 / 64.30
& 60.10 / 57.70 \\

TrackAny3D~\cite{wang2025trackany3d}
& 54.57 / 66.25
& 59.30 / 66.46
& 40.37 / 68.70
& 62.70 / 62.80
& 66.12 / 59.20
& \underline{61.01} / \underline{58.02} \\

P2P-voxel$^\dagger$~\cite{p2p}
& \textbf{59.22} / \underline{71.19}
& \textbf{64.61} / \textbf{71.98}
& \underline{45.64} / \underline{74.62}
& 64.42 / 65.37
& \underline{70.23} / \underline{66.08}
& 58.54 / 56.13 \\ 
\midrule

P2P-point~\cite{p2p}
& 55.92 / 66.64
& 62.14 / 68.45
& 39.68 / 65.59
& 62.50 / 63.44
& 69.04 / 65.14
& 57.90 / 55.46 \\

\rowcolor{myblue!18}
\textbf{SAVTrack (Ours)}
& 58.44 / 69.82
& \underline{64.27} / \underline{71.72}
& 42.54 / 68.31
& \textbf{65.16} / \textbf{66.18}
& \textbf{72.60} / \textbf{70.48}
& \textbf{64.02} / \textbf{61.67} \\

\bottomrule[0.4mm]
\end{tabular}}
\label{tab:NuScenes}
\vspace{-2mm}
\end{table*}

\begin{figure*}[!t]
\centering
\includegraphics[width=0.95\linewidth]{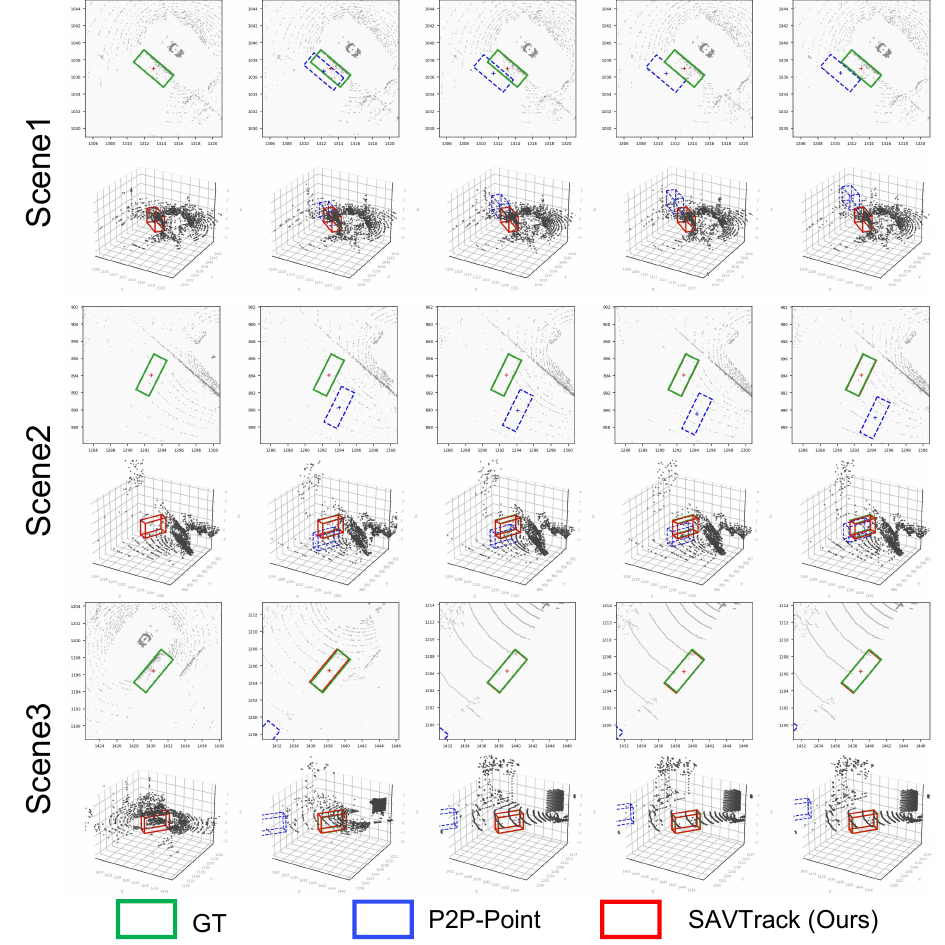}
\caption{Qualitative comparison of tracking results on nuScenes. \textcolor{green}{Green}, \textcolor{blue}{blue}, and \textcolor{red}{red} boxes denote the ground truth, P2P-point baseline~\cite{p2p}, and SAVTrack, respectively. SAVTrack provides more accurate
target localization across different target configurations and observation conditions.}
\label{vis_fig}
\vspace{-4mm}
\end{figure*}

\begin{figure*}[!t]
\centering
\includegraphics[width=0.8\textwidth]{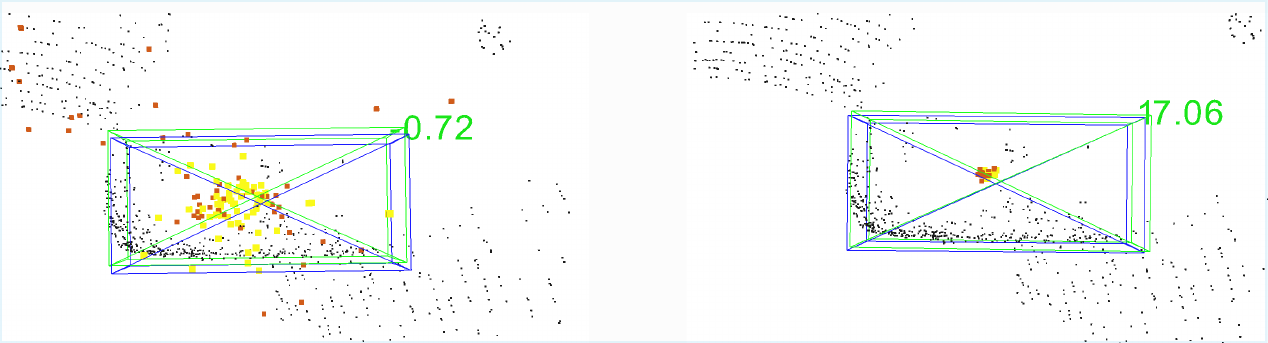}
\vspace{-3mm}
\caption{Visualization of voting results \textbf{without} (left) and \textbf{with} (right) SAV. With SAV, votes concentrate more tightly around the true object center, as low-confidence votes are excluded before aggregation.}
\label{vis_vote}
\vspace{-3mm}
\end{figure*}

\noindent\textbf{Comparison on nuScenes.}
We further evaluate SAVTrack on the nuScenes benchmark~\cite{nuScenes}, which provides a substantially larger and sparser tracking setting than KITTI, with 117,278 evaluation frames and point clouds captured by a 32-beam LiDAR. As shown in Tab.~\ref{tab:NuScenes}, SAVTrack achieves a mean Success/Precision of 58.44\%/69.82\%. Compared with the P2P-point~\cite{p2p}, SAV improves the mean Success and Precision by +2.52 and +3.18 points, respectively. The improvement is consistent across all five evaluated categories, including Car (+2.13/+3.27), Pedestrian (+2.86/+2.72), Truck (+2.66/+2.74), Trailer (+3.56/+5.34), and Bus (+6.12/+6.21). These consistent gains indicate that selectively suppressing unreliable center hypotheses benefits different target categories rather than a specific object type.

Although SAVTrack does not achieve the highest performance, it remains competitive with recent strong trackers and exhibits strong category-wise performance. Specifically, SAVTrack achieves the best Success/Precision on Truck, Trailer, and Bus, while obtaining the second-best results on Car. Compared with MBPTrack~\cite{mbptrack}, SAVTrack improves mean Success from 57.48\% to 58.44\% while achieving nearly identical mean Precision (69.82\% vs.\ 69.88\%). It also substantially outperforms M$^2$Track~\cite{m2track} and MoCUT~\cite{cutrack} in mean tracking accuracy. Notably, the gain over the voting baseline is larger on several challenging vehicle categories, particularly Trailer and Bus. Since category-level statistics jointly reflect point density, object scale, and geometric structure, they alone cannot determine whether the improvement originates from observation sparsity. We therefore provide a point-count-stratified analysis in Sec.~\ref{sec:sparsity_analysis} to examine the effect of sparsity more directly.

\subsection{Visualization}
\label{sec:visualization}

\noindent\textbf{Tracking Results.}
Fig.~\ref{vis_fig} presents qualitative tracking results on nuScenes~\cite{nuScenes} dataset. Compared with P2P-point, SAVTrack produces bounding boxes that are more consistently aligned with the ground-truth target across different scenes,
particularly under partial and sparse observations.

\noindent\textbf{Voting Results.}
Fig.~\ref{vis_vote} further visualizes the center hypotheses before
proposal aggregation. Dense voting produces a relatively dispersed vote
distribution, whereas SAV removes low-confidence hypotheses and yields a
more concentrated set of candidate centers around the target. This
qualitatively supports the motivation of reliability-aware pre-aggregation
selection.

\begin{table*}[!t]
\caption{Relationship between posterior confidence, vote localization quality, 
and local geometry on KITTI Car. Candidate votes are partitioned into five 
equal-frequency bins according to $p_j^i$. Planarity and curvature are computed from the local neighborhood of each seed.}
\centering
\resizebox{0.85\linewidth}{!}{%
\normalsize
\begin{tabular}{l|cccccc}
\toprule[0.4mm]
\textbf{Confidence bin} & \textbf{Mean conf.} & \textbf{Mean err.\ (m)}$\downarrow$ & \textbf{Med.\ err.\ (m)}$\downarrow$ & \textbf{Succ.\ ratio}$\uparrow$ & \textbf{Planarity}$\downarrow$ & \textbf{Curvature}$\uparrow$ \\
\midrule
Lowest 20\%  & 0.12 & 0.72 & 0.63 & 28.3\% & 0.81 & 0.042 \\
20--40\%     & 0.28 & 0.54 & 0.46 & 44.7\% & 0.74 & 0.061 \\
40--60\%     & 0.45 & 0.38 & 0.31 & 62.1\% & 0.66 & 0.083 \\
60--80\%     & 0.63 & 0.24 & 0.19 & 78.6\% & 0.57 & 0.108 \\
Highest 20\% & 0.84 & 0.13 & 0.10 & 89.4\% & 0.49 & 0.131 \\
\bottomrule[0.4mm]
\end{tabular}}
\label{tab:calibration}
\vspace{-4mm}
\end{table*}

\begin{table}[!t]
\caption{Ablation of vote aggregation strategies on KITTI. Mean denotes the overall result across all four KITTI categories.}
\centering
\resizebox{\linewidth}{!}{
\normalsize
\begin{tabular}{l|ccc|c}
\toprule[0.4mm]
\textbf{Strategy} & \textbf{Car} & \textbf{Ped.} & \textbf{Mean} & \textbf{Votes} \\
\midrule
(A) Dense        & 68.8 / 81.7 & 62.7 / 89.1 & 65.7 / 85.9  & $\approx$1536 \\
(B) Soft         & 69.2 / 82.2 & 63.1 / 89.5 & 66.0 / 86.3  & $\approx$1536 \\
(C) Top-1        & 69.5 / 82.8 & 63.5 / 90.0 & 66.4 / 86.8  & 128 \\
\midrule
\rowcolor{myblue!18}
\textbf{(D) SAVTrack}
             & \textbf{71.1 / 83.9}
             & \textbf{65.1 / 91.8}
             & \textbf{68.4 / 87.4}
             & $\approx$250 \\
\bottomrule[0.4mm]
\end{tabular}
}
\label{tab:ablation_vote_selection}
\vspace{-5mm}
\end{table}

\vspace{-2mm}
\subsection{Ablation Study}
\label{sec:ablation}

\noindent\textbf{Ablation on Vote Selection and Aggregation.}
Tab.~\ref{tab:ablation_vote_selection} compares different vote aggregation strategies under the same feature encoder, region-specific regressors, and proposal-generation pipeline. Soft posterior weighting (B) slightly improves over dense voting (A), increasing the overall Success/Precision from 65.7/85.9 to 66.0/86.3, indicating that the learned posterior provides useful confidence information but cannot fully suppress unreliable hypotheses. Top-1 selection (C) further improves performance to 66.4/86.8 by retaining only the highest-posterior vote per seed, although such aggressive selection may discard additional plausible hypotheses. In contrast, SAV (D) retains approximately 250 high-confidence votes and achieves the best overall result of 68.4/87.4, outperforming Top-1 by 1.6/1.1 on Car and 1.6/1.8 on Pedestrian. These results show that SAV benefits from selectively removing unreliable votes while preserving informative candidates, rather than simply minimizing the number of votes.

\begin{table}[!t]
\caption{Component analysis on KITTI. ``M-Hyp.'' denotes region-specific vote regression (one regressor per sub-region); ``Filter.'' denotes posterior-based confidence filtering before proposal aggregation.}
\centering
\resizebox{\linewidth}{!}{
\small
\begin{tabular}{l|cc|ccc|c}
\toprule[0.4mm]
\textbf{Variant}
& \textbf{M-Hyp.}
& \textbf{Filter}
& \textbf{Car}
& \textbf{Ped.}
& \textbf{Mean}
& \textbf{Votes} \\
\midrule
Baseline
& -- & --
& 68.8 / 81.7
& 62.7 / 89.1
& 65.7 / 85.9
& $\approx$1536 \\

M-Hyp. only
& \checkmark & --
& 69.6 / 82.4
& 63.3 / 90.2
& 67.2 / 87.0
& $\approx$1536 \\

\rowcolor{myblue!18}
\textbf{SAVTrack}
& \checkmark & \checkmark
& \textbf{71.1 / 83.9}
& \textbf{65.1 / 91.8}
& \textbf{68.4 / 87.4}
& $\approx$250 \\
\bottomrule[0.4mm]
\end{tabular}
}
\vspace{-2mm}
\label{tab:component_ablation}
\end{table}

\noindent\textbf{Component Analysis.} 
Tab.~\ref{tab:component_ablation} evaluates the contribution of region-specific multi-hypothesis regression and confidence filtering. Introducing multi-hypothesis regression alone improves the overall Success/Precision from 65.7/85.9 to 67.2/87.0, showing that region-specific regressors provide more flexible center hypotheses but still suffer from indiscriminate aggregation. Adding confidence filtering further raises the mean performance to 68.4/87.4 while reducing the retained votes from approximately 1,536 to 250. Similar gains are observed on both Car and Pedestrian, indicating that the main benefit of SAV comes from suppressing unreliable hypotheses before proposal clustering rather than from multi-hypothesis regression alone. The calibration and geometric characteristics of the learned confidence are further analyzed in \ref{analysis}.

\begin{table}[!t]
\caption{Sensitivity to the confidence threshold $\tau$ on KITTI. 
Mean denotes the overall result across all KITTI categories.}
\centering
\resizebox{\linewidth}{!}{
\normalsize
\begin{tabular}{c|ccc|c}
\toprule[0.4mm]
$\boldsymbol{\tau}$ & \textbf{Car} & \textbf{Ped.} & \textbf{Mean} & \textbf{Avg. votes} \\
\midrule
0.0 (w/o filter) & 69.6 / 82.4 & 63.3 / 90.2 & 67.2 / 87.0& $\approx$1536 \\
0.1 & 70.2 / 83.0 & 64.4 / 90.7 & 67.5 / 87.7 & $\approx$820 \\
\rowcolor{myblue!18}\textbf{0.3} & \textbf{71.1} / \textbf{83.9} & \textbf{65.1} / \textbf{91.8} & \textbf{68.4}/ \textbf{87.4} &\textbf{$\approx$250} \\
0.5 & 70.5 / 83.2 & 64.7 / 91.2 & 68.1/ 87.2 & $\approx$80 \\
\bottomrule[0.4mm]
\end{tabular}}
\label{tab:threshold_sensitivity}
\end{table}

\noindent\textbf{Confidence Threshold Sensitivity}
Tab.~\ref{tab:threshold_sensitivity} studies the effect of the gating threshold $\tau$. Increasing $\tau$ from 0 to 0.3 progressively removes low-confidence hypotheses and improves the overall tracking performance, with $\tau=0.3$ achieving the best result of 68.4/87.4 using approximately 250 votes. A larger threshold of 0.5 further reduces the vote set but slightly degrades performance, suggesting that overly aggressive filtering may discard informative hypotheses. We therefore use $\tau=0.3$ as the default setting.

\begin{table}[h]
\caption{Effect of the number of sub-regions $N_R$ on KITTI Car.}
\centering
\resizebox{0.75\linewidth}{!}{%
\footnotesize
\begin{tabular}{c|cc|c}
\toprule[0.4mm]
$\boldsymbol{N_R}$ & \textbf{Success} & \textbf{Precision} & \textbf{Avg. votes} \\
\midrule
4  & 69.8 & 82.6 & $\approx$185 \\
8  & 70.6 & 83.4 & $\approx$221 \\
\rowcolor{myblue!18}\textbf{12} & \textbf{71.1} & \textbf{83.9} & \textbf{$\approx$250} \\
16 & 70.9 & 83.6 & $\approx$276 \\
24 & 70.3 & 83.0 & $\approx$307 \\
\bottomrule[0.4mm]
\end{tabular}}
\label{tab:nr_ablation}
\vspace{-3mm}
\end{table}

\noindent\textbf{Number of Sub-regions.}
We study the sensitivity to the number of center-relative sub-regions $N_R$ in Tab.~\ref{tab:nr_ablation}. A small $N_R$ provides only a coarse discretization of the center-relative space, whereas an overly large $N_R$ increases the classification granularity and may leave fewer training samples for each region. Performance improves as $N_R$ increases from 4 to 12, reaching 71.1/83.9 Success/Precision, and then slightly decreases for larger values. We therefore set $N_R=12$ by default, which provides a good trade-off between representation granularity and learning stability.

\vspace{-2mm}
\subsection{More Analysis}
\label{analysis}

\noindent\textbf{Vote Confidence and Reliability Analysis.}
Tracking improvements alone do not reflect whether the learned posterior identifies reliable center hypotheses. We therefore analyze each candidate vote $v_j^i$ using its posterior confidence $p_j^i$ and localization error $e_j^i=\|\hat{\mathbf d}_j^i-\mathbf c\|_2$. As shown in Tab.~\ref{tab:calibration}, localization quality improves monotonically with confidence: the mean center error decreases from 0.72\,m in the lowest-confidence quintile to 0.13\,m in the highest, while the successful-vote ratio increases from 28.3\% to 89.4\%. Meanwhile, higher-confidence votes exhibit lower planarity and larger curvature, suggesting that the learned confidence is also associated with more distinctive local geometry. These results support the use of the sub-region posterior as a reliability proxy for vote selection.

\begin{table}[!t]
\caption{Spearman rank correlations between local geometric descriptors and
seed-level confidence / vote error on KITTI Car. Confidence is defined as the
maximum sub-region posterior of each seed. Correlations with
$|\rho|\geq0.3$ are highlighted in bold.}
\centering
\resizebox{\linewidth}{!}{%
\normalsize
\begin{tabular}{l|cc}
\toprule[0.4mm]
\textbf{Geometric descriptor} & \textbf{vs.\ Confidence} & \textbf{vs.\ Vote error} \\
\midrule
Planarity ($\uparrow$ = more planar)     & $\mathbf{-0.52}$ & $\mathbf{+0.47}$ \\
Curvature ($\uparrow$ = more curved)     & $\mathbf{+0.49}$ & $\mathbf{-0.43}$ \\
Normal variation ($\uparrow$ = more varied) & $\mathbf{+0.44}$ & $\mathbf{-0.38}$ \\
Geometric entropy ($\uparrow$ = more isotropic) & $\mathbf{+0.38}$ & $\mathbf{-0.34}$ \\
\bottomrule[0.4mm]
\end{tabular}}
\label{tab:correlation}
\end{table}

\noindent\textbf{Geometric Correlation Analysis.}
To examine whether the learned confidence is associated with local geometric
structure, we compute planarity, curvature, normal variation, and geometric
entropy from the $k$-NN neighborhood of each seed. For each seed, we use the
maximum sub-region posterior as its confidence and the localization error of
the corresponding highest-confidence vote. Tab.~\ref{tab:correlation} reports
their Spearman rank correlations.

Planarity is negatively correlated with confidence ($\rho=-0.52$) and
positively correlated with vote error ($\rho=+0.47$), suggesting that locally
planar regions tend to provide weaker localization evidence. In contrast,
curvature, normal variation, and geometric entropy are positively correlated
with confidence and negatively correlated with error. These trends are
consistent with the confidence-bin analysis in Tab.~\ref{tab:calibration} and
support an association between learned confidence and richer local geometric
variation, which is consistent with SAV estimating reliability jointly from local point-wise
evidence and inter-frame motion context.

\begin{table}[h]
\caption{Computational overhead. Speed is measured with batch size~1 on a single RTX~3090.}
\centering
\resizebox{0.7\linewidth}{!}{%
\footnotesize
\begin{tabular}{l|cc}
\toprule[0.4mm]
\textbf{Model} & \textbf{Params (M)} & \textbf{FPS} \\
\midrule
Baseline & 7.39 & 98 \\
\rowcolor{myblue!18}\textbf{SAVTrack (Ours)} & \textbf{7.76} & \textbf{82} \\
\bottomrule[0.4mm]
\end{tabular}}
\label{tab:overhead}
\vspace{-3mm}
\end{table}

\noindent\textbf{Computational Overhead.} Tab.~\ref{tab:overhead} shows that the SAV module adds only 0.4M parameters while maintaining real-time inference at 82~FPS, demonstrating that the proposed gating mechanism is lightweight and practical within this tracking pipeline.

\begin{table}[h]
\caption{Per-category improvement (Success/Precision) over the baseline on nuScenes. $\Delta$S and $\Delta$P denote absolute gains in Success and Precision; $\overline{N}_{\text{pts}}$ is the average number of foreground points per target.}
\footnotesize 
\centering
\setlength{\tabcolsep}{4pt} 
\begin{tabular}{l|c|cc|cc}
\toprule[0.4mm]
\textbf{Category} & $\overline{N}_{\textbf{pts}}$ & \textbf{Baseline} & \textbf{SAVTrack} & $\Delta$\textbf{S} & $\Delta$\textbf{P} \\
\midrule
Car        & 312 & 62.14 / 68.45 & \textbf{64.27 / 71.72} & \textbf{+2.13} & \textbf{+3.27} \\
Ped. & 87  & 39.68 / 65.59 & \textbf{42.54 / 68.31} & \textbf{+2.86} & \textbf{+2.72} \\
Truck      & 275 & 62.50 / 63.44 & \textbf{65.16 / 66.18} & \textbf{+2.66} & \textbf{+2.74} \\
Trailer    & 348 & 69.04 / 65.14 & \textbf{72.60 / 70.48} & \textbf{+3.56} & \textbf{+5.34} \\
Bus        & 375 & 57.90 / 55.46 & \textbf{64.02 / 61.67} & \textbf{+6.12} & \textbf{+6.21} \\
\bottomrule[0.4mm]
\end{tabular}
\label{tab:category_gain}
\vspace{-2mm}
\end{table}

\noindent\textbf{Per-category Analysis.}
Tab.~\ref{tab:category_gain} reports the category-wise improvements over the baseline on nuScenes~\cite{nuScenes}. SAVTrack consistently improves both Success and Precision across all five categories, with gains ranging from +2.13/+3.27 on Car to +6.12/+6.21 on Bus. Notably, the improvement does not monotonically correlate with the average number of foreground points: although Pedestrian has the sparsest observations, Bus and Trailer exhibit larger gains despite having considerably more points. This suggests that category-level improvements are influenced not only by point density, but also by factors such as object geometry and observation structure. We therefore further isolate the effect of sparsity through the point-count-stratified analysis in Sec.~\ref{sec:sparsity_analysis}.

\begin{table*}[!t]
\caption{Sparsity-stratified analysis on nuScenes. Frames are grouped by the number of foreground points in the ground-truth box. $\Delta$S and $\Delta$P denote the absolute gain of SAVTrack over the voting baseline; ``Votes kept'' is the average number of votes retained after hard gating in each bin.}
\centering
\resizebox{0.95\textwidth}{!}{%
\normalsize
\begin{tabular}{c|cc|cc|cc|cc|c}
\toprule[0.4mm]
\multirow{2}{*}{\textbf{Points / target}} & \multicolumn{2}{c|}{\textbf{Voting baseline}} & \multicolumn{2}{c|}{\textbf{Multi-hyp.\ only}} & \multicolumn{2}{c|}{\textbf{SAVTrack (Ours)}} & \multicolumn{2}{c|}{\textbf{$\Delta$}} & \multirow{2}{*}{\textbf{Avg.\ votes kept}} \\
 & S & P & S & P & S & P & $\Delta$S & $\Delta$P & \\
\midrule
0--10   & 31.2 & 40.8 & 32.5 & 42.1 & 36.4 & 47.2 & \textbf{+5.2} & \textbf{+6.4} & $\approx$62 \\
10--20  & 40.1 & 50.3 & 41.3 & 51.8 & 44.5 & 56.1 & \textbf{+4.4} & \textbf{+5.8} & $\approx$115 \\
20--30  & 47.3 & 57.2 & 48.6 & 58.9 & 51.2 & 62.4 & +3.9 & +5.2 & $\approx$165 \\
30--40  & 53.1 & 62.8 & 54.2 & 64.0 & 56.3 & 66.2 & +3.2 & +3.4 & $\approx$210 \\
40--50  & 57.4 & 66.5 & 58.3 & 67.6 & 59.8 & 69.1 & +2.4 & +2.6 & $\approx$248 \\
$>$50   & 63.2 & 72.8 & 64.1 & 73.7 & 65.5 & 75.0 & +2.3 & +2.2 & $\approx$295 \\
\bottomrule[0.4mm]
\end{tabular}}
\label{tab:sparsity_analysis}
\end{table*}

\begin{table*}[!htbp]
\centering
\caption{Ablation of static and motion-conditioned reliability estimation on KITTI. ``Local'' and ``Motion'' indicate the cues used by the sub-region posterior estimator; all variants otherwise share the same motion-centric tracking backbone and proposal-generation pipeline.}
\label{tab:spot_comparison}
\small
\setlength{\tabcolsep}{5pt}
\begin{tabular}{lccccc}
\toprule
\textbf{Variant}
& \textbf{Local Cue.}
& \textbf{Motion Cue.}
& \textbf{Selection}
& \textbf{Car}
& \textbf{Ped} \\
\midrule
Dense baseline
& -- 
& --
& Dense
& 68.8 / 81.7
& 62.7 / 89.1 \\

Static-only reliability~\cite{du2020spot}
& $\checkmark$
& $\times$
& Hard
& 69.1 / 82.0
& 62.8 / 89.4 \\

Motion-aware + soft weighting
& $\checkmark$
& $\checkmark$
& Soft
& 69.2 / 82.2
& 63.1 / 89.5 \\

\textbf{SAVTrack}
& $\checkmark$
& $\checkmark$
& \textbf{Hard}
& \textbf{71.1 / 83.9}
& \textbf{65.1 / 91.8} \\
\bottomrule
\end{tabular}
\vspace{-2mm}
\end{table*}

\subsection{Sparsity-Stratified Analysis}
\label{sec:sparsity_analysis}

To isolate the effect of observation sparsity from category-specific confounders (size, shape, typical occlusion), we stratify test frames by the number of foreground points on the target, following the protocol of P2P~\cite{p2p}. For each frame, we count the points inside the ground-truth bounding box and assign the frame to one of six bins: 0--10, 10--20, 20--30, 30--40, 40--50, and $>$50 points. We then compute Success and Precision separately for each bin, comparing the baseline, the multi-hypothesis-only variant, and SAVTrack. Our diagnostic hypothesis is that the benefit of reliability-aware gating
increases as observations become sparser, since unreliable hypotheses may have a larger impact on proposal formation when only limited target evidence is available.

\noindent\textbf{Sparsity Trends.}
Tab.~\ref{tab:sparsity_analysis} shows a clear dependence on target point density. While all methods degrade as observations become sparser, the gain of SAVTrack over the voting baseline increases consistently: $\Delta$S grows from +2.3 in the $>50$-point bin to +5.2 in the 0--10-point bin, while $\Delta$P increases from +2.2 to +6.4. Multi-hypothesis regression alone yields only modest and relatively stable gains across bins, suggesting that the stronger improvement under sparse observations primarily comes from reliability-aware filtering. The retained vote set also becomes substantially smaller in sparse bins, indicating more selective hypothesis retention under limited observations. These results support the hypothesis that reliability-aware gating is particularly beneficial when target observations are sparse.

\paragraph{Effect of Motion-Conditioned Reliability Estimation.}
To examine whether static selective voting is sufficient for temporal 3D tracking, we construct a SPOT-inspired variant in which the sub-region posterior is estimated using only the current-frame seed feature, while all other components remain unchanged. As shown in Tab.~\ref{tab:spot_comparison}, the static-only variant reaches 69.1/82.0 on Car and 62.8/89.4 on Pedestrian, whereas incorporating inter-frame motion cues improves the results to 71.1/83.9 and 65.1/91.8, respectively. This suggests that temporal motion context provides complementary information for assessing vote reliability. Moreover, under the same motion-aware representation, hard gating clearly outperforms soft weighting, improving Car by 1.9/1.7 and Pedestrian by 2.0/2.3 points. These results indicate that SAV benefits from both motion-conditioned reliability estimation and the explicit removal of low-confidence hypotheses before proposal clustering.

\section{Conclusion}
\label{sec:conclusion}

In this work, we investigate heterogeneous vote reliability in voting-based 3D single object tracking, where unreliable seed-to-center hypotheses can interfere with proposal formation when aggregated indiscriminately. To address this issue, we propose SAVTrack with Selective Vote Aggregation (SAV), a lightweight pre-aggregation mechanism that predicts a center-relative sub-region posterior from joint point-wise and inter-frame motion representations and uses the resulting confidence to filter unreliable candidate votes before clustering. Experiments on KITTI and nuScenes show that SAVTrack consistently improves the corresponding voting baseline while maintaining real-time inference at 82 FPS. Extensive analyses further show that the learned confidence is predictive of vote localization quality, correlates with local geometric structure, and provides larger gains under sparse target observations. Controlled comparisons also demonstrate the complementary benefits of motion-conditioned reliability estimation and hard pre-aggregation gating. These results highlight the importance of explicitly considering hypothesis reliability before proposal aggregation in point-cloud tracking.

\noindent\textbf{Limitations and Future Work.}
SAV currently relies on a fixed, uniformly defined center-relative partition, which may not fully capture anisotropic or object-dependent center distributions. Learning adaptive partitions or adopting hierarchical coarse-to-fine voting could provide more flexible hypothesis modeling. In addition, reliability is estimated independently for individual seeds, without explicitly considering geometric consistency among retained hypotheses. Incorporating set-level consensus or temporal consistency verification may further improve robustness under severe sparsity and occlusion. Future work may also explore reliability-aware hypothesis selection in other point-cloud localization tasks where candidate predictions exhibit heterogeneous quality.

\section*{CrediT authorship contribution statement}
\textbf{Sifan Zhou:} Investigation, Methodology, Software, Validation, Writing - original draft, Writing - review \& editing, Project administration. \textbf{Linyue Tan:} Methodology, Validation, Visualization, Writing - original draft, Writing - review \& editing. \textbf{Qiwei Wang:} Software, Formal analysis, Validation, Writing -original draft \& editing. \textbf{Ziyu Zhao:} Formal analysis, Writing - review \& editing. \textbf{Xiaobo Lu:} Methodology, Writing - review, Project administration, Funding acquisition.

\section*{Declaration of competing interest}

The authors declare that they have no known competing financial interests or personal relationships that could have appeared to influence the work reported in this paper.

\section*{Acknowledgements}
This work was supported by the National Natural Science Foundation of China (No. 62271143), the Frontier Technologies R\&D Program of Jiangsu (No. BF2024060). On computing resources, this work was supported by the Big Data Computing Center of Southeast University.

{\small
\bibliographystyle{cas-model2-names}
\bibliography{egbib}
}

\end{document}